\documentclass[runningheads]{llncs}
\usepackage[mobile]{eccv}

\usepackage{eccvabbrv}
\usepackage{graphicx}
\usepackage{booktabs}
\usepackage{wrapfig}
\usepackage{tikz}
\usepackage{colortbl}
\usepackage{fp}
\usepackage{microtype}
\usepackage[accsupp]{axessibility}
\usepackage{threeparttable}

\usepackage{silence}
\usepackage{soul}
\setuldepth{foobar}

\newcommand{\up}{\,\raisebox{0.4ex}{\textcolor{black!60}{$\scriptstyle\blacktriangle$}}}
\newcommand{\down}{\,\raisebox{0.4ex}{\textcolor{black!60}{$\scriptstyle\blacktriangledown$}}}

\newcommand{\inlineheading}[1]{\vspace{1mm}\noindent\textbf{{#1}.}\hspace{0.5em}}

\newcommand{\fpmin}[6]{%
  \FPmin{\result}{#2}{#3}%
  \FPmin\result{\result}{#4}%
  \FPmin\result{\result}{#5}%
  \FPmin{#1}{\result}{#6}%
}
\newcommand{\fpmax}[6]{%
  \FPmax{\result}{#2}{#3}%
  \FPmax\result{\result}{#4}%
  \FPmax\result{\result}{#5}%
  \FPmax{#1}{\result}{#6}%
}

\def\fpnorm#1#2#3{%
\newcommand{#1}[1]{%
  \FPeval{\blendfactor}{(##1 - #2)/(#3 - #2)}%
  \FPeval{\xlt}{(\blendfactor * 2.0)}%
  \FPeval{\xgt}{(\blendfactor * 2.0) - 1.0}%
  \FPeval{\ylt}{(0.75)}%
  \FPeval{\ygt}{(1.0 - 0.75)}%
  \FPeval{\biaslt}{(\xlt / ((((1.0/\ylt) - 2.0)*(1.0 - \xlt))+1.0)) * 0.5}%
  \FPeval{\biasgt}{(\xgt / ((((1.0/\ygt) - 2.0)*(1.0 - \xgt))+1.0)) * 0.5 + 0.5}%
  \FPeval{\t}{\blendfactor}%
  \FPeval{\xt}{0.5}%
  \FPiflt\t\xt \FPset{\blendfactor}{\biaslt} \else \FPset{\blendfactor}{\biasgt} \fi%

  \FPeval{\blendred}{1-(\blendfactor-0.05)/(1.0-0.05)}%
  \FPmin{\blendred}{\blendred}{1.0}%
  \FPmax{\blendred}{\blendred}{0.0}%
  \FPset{\blendgreen}{\blendred}%
  \FPset{\blendblue}{1.0}%
  \FPset{\blendalpha}{0.25}%
  \FPeval{\blendblue}{ \blendalpha * \blendblue  + (1 - \blendalpha)}%
  \FPeval{\blendgreen}{\blendalpha * \blendgreen + (1 - \blendalpha)}%
  \FPeval{\blendred}{  \blendalpha * \blendred   + (1 - \blendalpha)}%
  \xdef\temp{\noexpand\cellcolor[rgb]{\blendred, \blendgreen, \blendblue}{##1}}
  \temp
  }%
}

\def\fpnorminv#1#2#3{%
\newcommand{#1}[1]{%
  \FPeval{\blendfactor}{1 - (##1 - #2)/(#3 - #2)}%
  \FPeval{\xlt}{(\blendfactor * 2.0)}%
  \FPeval{\xgt}{(\blendfactor * 2.0) - 1.0}%
  \FPeval{\ylt}{(0.75)}%
  \FPeval{\ygt}{(1.0 - 0.75)}%
  \FPeval{\biaslt}{(\xlt / ((((1.0/\ylt) - 2.0)*(1.0 - \xlt))+1.0)) * 0.5}%
  \FPeval{\biasgt}{(\xgt / ((((1.0/\ygt) - 2.0)*(1.0 - \xgt))+1.0)) * 0.5 + 0.5}%
  \FPeval{\t}{\blendfactor}%
  \FPeval{\xt}{0.5}%
  \FPiflt\t\xt \FPset{\blendfactor}{\biaslt} \else \FPset{\blendfactor}{\biasgt} \fi%

  \FPeval{\blendred}{1-(\blendfactor-0.05)/(1.0-0.05)}%
  \FPmin{\blendred}{\blendred}{1.0}%
  \FPmax{\blendred}{\blendred}{0.0}%
  \FPset{\blendgreen}{\blendred}%
  \FPset{\blendblue}{1.0}%
  \FPset{\blendalpha}{0.25}%
  \FPeval{\blendblue}{ \blendalpha * \blendblue  + (1 - \blendalpha)}%
  \FPeval{\blendgreen}{\blendalpha * \blendgreen + (1 - \blendalpha)}%
  \FPeval{\blendred}{  \blendalpha * \blendred   + (1 - \blendalpha)}%
  \xdef\temp{\noexpand\cellcolor[rgb]{\blendred, \blendgreen, \blendblue}{##1}}
  \temp
  }%
}

\newcommand{\mynormcolor}[2]{%
\FPeval{\nc}{}%

}

\def\fpnormlog#1#2#3{%
\newcommand{#1}[1]{%
  \FPln{\logval}{##1}%
  \FPln{\logmin}{#2}%
  \FPln{\logmax}{#3}%
  \FPeval{\blendfactor}{(\logval - \logmin)/(\logmax - \logmin)}%
  \FPeval{\xlt}{(\blendfactor * 2.0)}%
  \FPeval{\xgt}{(\blendfactor * 2.0) - 1.0}%
  \FPeval{\ylt}{(0.9)}%
  \FPeval{\ygt}{(1.0 - 0.9)}%
  \FPeval{\biaslt}{(\xlt / ((((1.0/\ylt) - 2.0)*(1.0 - \xlt))+1.0)) * 0.5}%
  \FPeval{\biasgt}{(\xgt / ((((1.0/\ygt) - 2.0)*(1.0 - \xgt))+1.0)) * 0.5 + 0.5}%
  \FPeval{\t}{\blendfactor}%
  \FPeval{\xt}{0.5}%
  \FPiflt\t\xt \FPset{\blendfactor}{\biaslt} \else \FPset{\blendfactor}{\biasgt} \fi%
  \FPpow{\blendfactor}{\blendfactor}{2}%
  \FPeval{\blendred}{1-(\blendfactor-0.05)/(1.0-0.05)}%
  \FPmin{\blendred}{\blendred}{1.0}%
  \FPmax{\blendred}{\blendred}{0.0}%
  \FPset{\blendgreen}{\blendred}%
  \FPset{\blendblue}{1.0}%
  \FPset{\blendalpha}{0.25}%
  \FPeval{\blendblue}{ \blendalpha * \blendblue  + (1 - \blendalpha)}%
  \FPeval{\blendgreen}{\blendalpha * \blendgreen + (1 - \blendalpha)}%
  \FPeval{\blendred}{  \blendalpha * \blendred   + (1 - \blendalpha)}%
  \xdef\temp{\noexpand\cellcolor[rgb]{\blendred, \blendgreen, \blendblue}{##1}}
  \temp
  }%
}

\def\fpnorminvlog#1#2#3{%
\newcommand{#1}[1]{%
  \FPln{\logval}{##1}%
  \FPln{\logmin}{#2}%
  \FPln{\logmax}{#3}%
  \FPeval{\blendfactor}{1 - (\logval - \logmin)/(\logmax - \logmin)}%
  \FPeval{\xlt}{(\blendfactor * 2.0)}%
  \FPeval{\xgt}{(\blendfactor * 2.0) - 1.0}%
  \FPeval{\ylt}{(0.9)}%
  \FPeval{\ygt}{(1.0 - 0.9)}%
  \FPeval{\biaslt}{(\xlt / ((((1.0/\ylt) - 2.0)*(1.0 - \xlt))+1.0)) * 0.5}%
  \FPeval{\biasgt}{(\xgt / ((((1.0/\ygt) - 2.0)*(1.0 - \xgt))+1.0)) * 0.5 + 0.5}%
  \FPeval{\t}{\blendfactor}%
  \FPeval{\xt}{0.5}%
  \FPiflt\t\xt \FPset{\blendfactor}{\biaslt} \else \FPset{\blendfactor}{\biasgt} \fi%
  \FPpow{\blendfactor}{\blendfactor}{2}%
  \FPeval{\blendred}{1-(\blendfactor-0.05)/(1.0-0.05)}%
  \FPmin{\blendred}{\blendred}{1.0}%
  \FPmax{\blendred}{\blendred}{0.0}%
  \FPset{\blendgreen}{\blendred}%
  \FPset{\blendblue}{1.0}%
  \FPset{\blendalpha}{0.25}%
  \FPeval{\blendblue}{ \blendalpha * \blendblue  + (1 - \blendalpha)}%
  \FPeval{\blendgreen}{\blendalpha * \blendgreen + (1 - \blendalpha)}%
  \FPeval{\blendred}{  \blendalpha * \blendred   + (1 - \blendalpha)}%
  \xdef\temp{\noexpand\cellcolor[rgb]{\blendred, \blendgreen, \blendblue}{##1}}
  \temp
  }%
}

\usepackage{hyperref}
\usepackage{orcidlink}
\usepackage[sectionbib]{chapterbib}

\begin{document}
\title{Free-Range Gaussians: Non-Grid-Aligned Generative 3D Gaussian Reconstruction}

\titlerunning{Free-Range Gaussians}

\author{Akhmedkhan (Ahan) Shabanov\inst{1, 2} \and
Peter Hedman\inst{1} \and
Ethan Weber\inst{1} \and
Zhengqin Li\inst{1} \and
Denis Rozumny\inst{1} \and
Gael Le Lan\inst{1} \and
Naina Dhingra\inst{1} \and
Lei Luo\inst{1} \and
Andrea Vedaldi\inst{1, 3} \and
Christian Richardt\inst{1} \and
Andrea Tagliasacchi\inst{2, 4} \and
Bo Zhu\inst{1} \and
Numair Khan\inst{1}
}

\authorrunning{A.~Shabanov et al.}

\institute{Meta Reality Labs \and
Simon Fraser University \and
University of Oxford \and
University of Toronto
}

\maketitle

\begin{abstract}
We present \emph{Free-Range Gaussians}, a multi-view reconstruction method that predicts non-pixel, non-voxel-aligned 3D Gaussians from as few as four images.
This is done through flow matching over Gaussian parameters.
Our generative formulation of reconstruction allows the model to be supervised with non-grid-aligned 3D data, and enables it to synthesize plausible content in unobserved regions.
Thus, it improves on prior methods that produce highly redundant grid-aligned Gaussians, and suffer from holes or blurry conditional means in unobserved regions.
To handle the number of Gaussians needed for high-quality results, we introduce a hierarchical patching scheme to group spatially related Gaussians into joint transformer tokens, halving the sequence length while preserving structure.
We further propose a timestep-weighted rendering loss during training, and photometric gradient guidance and classifier-free guidance at inference to improve fidelity. 
Experiments on Objaverse and Google Scanned Objects show consistent improvements over pixel and voxel-aligned methods while using significantly fewer Gaussians, with large gains when input views leave parts of the object unobserved.
\keywords{3D Gaussian Splatting \and Sparse-View Reconstruction \and Flow Matching \and Generative 3D}
\end{abstract}

\section{Introduction}%
\label{sec:intro}

Reconstructing 3D objects from sparse images is a long-standing challenge in computer vision.
3D Gaussian splatting (3DGS) \cite{KerblKLD2023} has emerged as a compelling representation offering real-time differentiable rendering with high visual quality.
This has spurred feed-forward methods that predict 3DGS from sparse views in a single forward pass, bypassing costly per-scene optimization.

The dominant approach is \emph{pixel-aligned} reconstruction: each predicted Gaussian is anchored to a pixel in one of the input views.
Methods such as Splatter Image \cite{SzymaRV2024}, DepthSplat \cite{XuPWBBGP2025}, LGM \cite{TangCCWZL2024}, and GS-LRM \cite{ZhangBTXZSX2024} all produce Gaussians that inherit the spatial structure of the input images.
While effective in observed regions, the resulting representation is highly redundant, and suffers from holes and artifacts in unobserved regions (\cref{fig:non-pixel-aligned}).
A common line of work addresses this problem by synthesizing novel views to observe hidden parts, and then reconstructing from the expanded set of images \cite{LiuWHTZV2023, LiuLZLLKW2024, SzymaZSGBHMBH2025}.
However, identifying the novel viewpoints presupposes unavailable geometric knowledge, and inconsistent synthesized views result in low-quality reconstruction.

\begin{figure}[t]
    \centering
    \includegraphics[trim=0cm 11.0cm 5cm 0cm, clip=true, width=\linewidth]{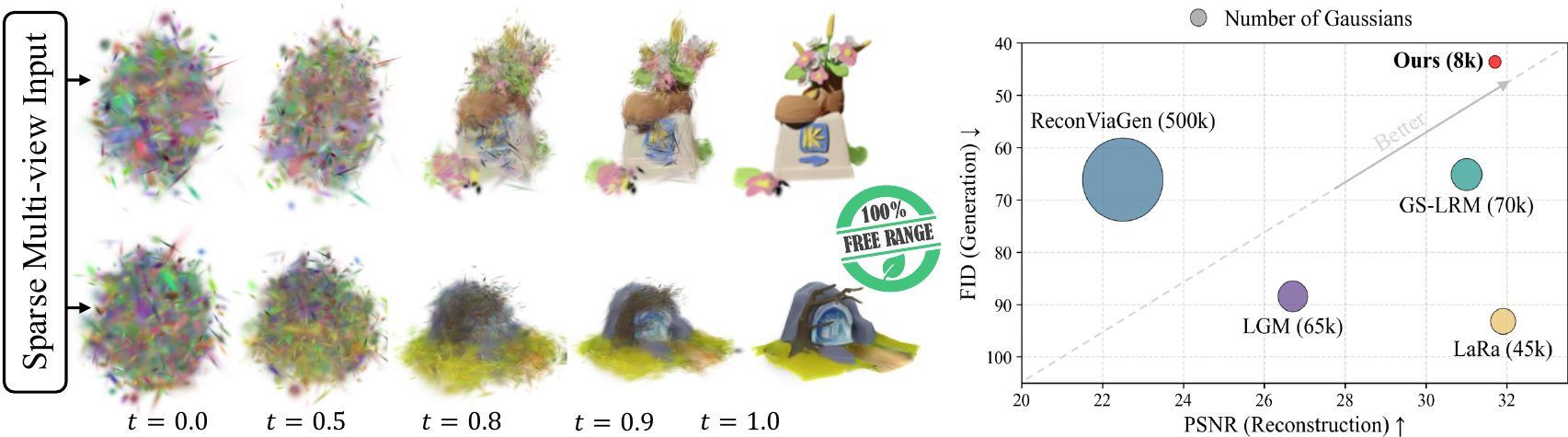}
    \caption{\label{fig:teaser}%
        \textbf{Non-grid-aligned generative reconstruction.}
        From four sparse views, our method produces a complete 3D Gaussian reconstruction via flow matching, achieving state-of-the-art quality for both observed (high PSNR) and unobserved regions (low FID) with ${\sim}10\times$ fewer Gaussians than grid-aligned methods.
    }
    \vspace{-1.5em}
\end{figure}

\begin{figure}[t]
\centering\includegraphics[trim=0cm 9cm 4.7cm 0cm, clip=true, width=0.9\linewidth]{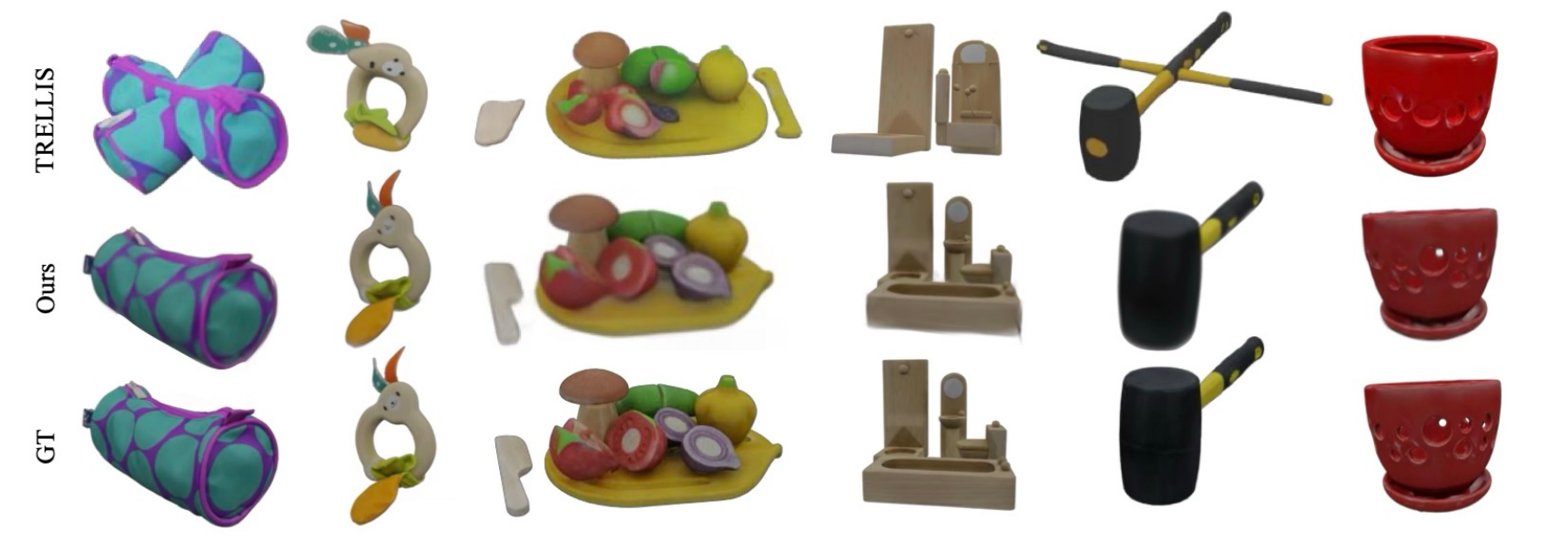}
\caption{\textbf{Generation vs. Reconstruction.}
State-of-the-art generative methods like TRELLIS \cite{XiangLXDWZCTY2025} produce sharp output but suffer from weak input conditioning as they fail to preserve object orientation, and hallucinate geometry and appearance even in observed regions.
Our results remain aligned with input-view poses, geometry, and appearance.
Please see the supplementary material for additional results.
}%
\label{fig:results-vs_generative}
\end{figure}

An alternative approach decouples the reconstruction from input pixels by using a volumetric grid \cite{ChenXETG2024, WangCZLWFQZCZ2025}.
However, the representation remains redundant, and detail is bound by grid resolution.
Further, trained only with a photometric loss, these methods predict blurry conditional means for unobserved areas.
Generative methods like TRELLIS \cite{XiangLXDWZCTY2025} hallucinate sharp content in unobserved areas, but are prone to ignore the conditioning images and poses (\cref{fig:results-vs_generative}).

We refer to the above-mentioned approaches as grid-aligned since both anchor Gaussians to a fixed spatial grid, whether 2D image pixels or 3D voxels.
We address their limitations through \emph{generative reconstruction}: instead of deterministically regressing Gaussian parameters, we use flow matching \cite{LipmaCBNL2023} to model the conditional distribution of 3D Gaussians given sparse views.
Thus, our model
iteratively denoises Gaussian parameters from noise.
As the output is not tied to pixels or voxels, the model can place Gaussians anywhere in 3D space, producing a more compact representation.
Furthermore, by sampling from the conditional distribution rather than regressing its mean, our method produces sharp completions for unseen regions while staying faithful to observed views.

A key challenge is scale: high-quality reconstructions require denoising tens of thousands of Gaussian tokens.
We construct a level-of-detail (LoD) hierarchy \cite{KerblMKWLD2024} over ground-truth Gaussians for coarse-to-fine training, and propose a patchification strategy that cuts the number of tokens in half while preserving locality.
To recover details and improve fidelity, we combine three complementary mechanisms:
(1)~a photometric loss that emphasizes details at later denoising stages,
(2)~photometric gradient guidance \cite{MuZGWLWXDYC2024} to steer denoising toward input views, and
(3)~classifier-free guidance \cite{HoS2021}.
In summary, our contributions are:
\vspace{-.5em}
\begin{itemize}
    \item A method to produce compact and complete non-grid-aligned 3D Gaussian reconstructions from sparse views via flow matching.
    \item A coarse-to-fine training and Gaussian patchification scheme based on a hierarchical LoD structure.
    \item Training and inference-time guidance mechanisms to recover high-frequency details and improve input fidelity.
\end{itemize}

\section{Related Work}%
\label{sec:related-work}

\inlineheading{Per-Pixel Feed-Forward 3D Reconstruction}
A large family of methods anchors each predicted 3D element to an input pixel or patch.
Splatter Image \cite{SzymaRV2024} predicts one Gaussian per pixel from a single view; LGM \cite{TangCCWZL2024} does so from four views using a U-Net.
pixelSplat \cite{CharaLTS2024} produces per-pixel Gaussians from stereo image pairs, while GRM \cite{XuSYCYPSW2024} scales the paradigm to high-resolution outputs via a large Gaussian reconstruction model.
GS-LRM \cite{ZhangBTXZSX2024} extends LRM \cite{HongZGBZLLSBT2024} with pixel-aligned transformer tokens and Plücker ray embeddings.
Long-LRM \cite{ZiwenTZBLHFX2025} further scales feed-forward Gaussian prediction to 32 high-resolution views by replacing the quadratic-cost transformer with a hybrid Mamba2--transformer architecture.
DepthSplat \cite{XuPWBBGP2025} and MVSplat \cite{ChenXZZPGCC2024} create Gaussians from multi-view stereo depth and cost-volume cues, respectively.
MVSNeRF \cite{ChenXZZXYS2021} was an early work combining plane-sweep cost volumes with volume rendering for generalizable radiance field reconstruction from multi-view stereo.
AnySplat \cite{JiangMXLRJXYPZLD2025} extends the paradigm of feed-forward geometry foundation models like DUST3R \cite{WangLCCR2024}, VGGT \cite{WangCKVRN2025} and MapAnything \cite{KeethMSPZFKZWALLBRRSK2026} to uncalibrated collections by jointly predicting Gaussians and cameras.
PF-LRM \cite{WangTBXLSWXZ2024} addresses the pose-free setting by jointly predicting shape and camera parameters, while Splatt3R \cite{SmartZLP2024} and MASt3R \cite{LeroyCR2024} extend DUST3R to predict Gaussians and dense stereo matches from uncalibrated image pairs.
These approaches are fast and accurate in observed regions, but cannot place Gaussians in unobserved regions, limiting them to partial reconstructions.

\inlineheading{Generate-and-Reconstruct Approaches}
To cover unobserved regions, several methods first synthesize novel views and then reconstruct from the expanded set.
Zero-1-to-3 \cite{LiuWHTZV2023} and SyncDreamer \cite{LiuLZLLKW2024} generate the expanded set via view-conditioned diffusion for downstream 3D reconstruction.
One-2-3-45++ \cite{LiuSCZXWCZGS2024} extends this pipeline with consistent multi-view generation and 3D diffusion, while InstantMesh \cite{XuCGWGS2024} streamlines it by coupling a multi-view diffusion model with a sparse-view large reconstruction model for efficient mesh extraction.
ReconFusion \cite{WuMHPGWSVBPH2024} regularizes a neural radiance field (NeRF) \cite{MildeSTBRN2020} at novel poses with a diffusion prior, but still requires per-scene optimization.
Similarly, iFusion \cite{WuCSYS2025} leverages Zero-1-to-3 predictions within an optimization pipeline to align poses and generate novel views, but the inconsistency between synthesized views still limits reconstruction quality.
Bolt3D \cite{SzymaZSGBHMBH2025} generates per-pixel Gaussian maps from a multi-view diffusion model and fuses them into a scene.
Wonderland \cite{LiangCGQKTPTR2025} and Lyra \cite{BahmaSRHJTTLGFLGR2026} both leverage video diffusion models to produce 3D Gaussians, with the latter using self-distillation without requiring multi-view training data.
These methods offer test-time adaptivity, but selecting the viewpoints to synthesize requires a priori scene understanding,
and artifacts from inconsistent generated images can propagate to the 3D output.

\inlineheading{Full Reconstruction via Structured Representations}
An alternative is to predict 3D content on a volumetric grid.
Among early work in this direction, pixelNeRF \cite{YuYTK2021} projects image features onto 3D query points.
More recently, LRM \cite{HongZGBZLLSBT2024} predicts tri-planes from a vision transformer, whereas LaRa \cite{ChenXETG2024} and VolSplat \cite{WangCZLWFQZCZ2025} predict Gaussians on voxel grids.
MeshLRM \cite{WeiZBTLDSSX2024} adapts the LRM architecture to directly produce high-quality meshes rather than radiance fields.
SCube \cite{RenLLWLCFWH2024} combines a hierarchical sparse-voxel scaffold with latent diffusion.
While decoupling the output from input pixels, the quality of these methods is bound by the grid resolution.
Further, under photometric losses, they predict the conditional mean of unobserved regions, yielding complete but blurry results.

\inlineheading{Generative 3D Models}
DreamFusion \cite{PooleJBM2023} distills 2D diffusion priors into 3D but requires per-object optimization.
Feed-forward alternatives apply diffusion or flow matching directly to 3D representations:
Point-E \cite{NichoJDMC2022} generates point clouds, and 3DShape2VecSet \cite{ZhangTNW2023} diffuses over neural-field latents.
Direct3D \cite{WuLZZXTCY2024} applies a 3D latent diffusion transformer for scalable image-to-3D generation.
TripoSG \cite{LiZLWLYLGLOC2026} employs large-scale rectified flow models over a point-cloud-structured latent space, achieving high-fidelity shape generation.
Among Gaussian-based generators, GaussianCube \cite{ZhangCYWZTCG2024} is tied to a voxel-structured representation, which inherits grid-capacity limits.
The same is true of TRELLIS \cite{XiangLXDWZCTY2025}, although it encodes the grid into a highly compact latent representation.
L3DG \cite{RoessMPBKDN2024} performs latent diffusion with a VQ-VAE \cite{OordVK2017}, but its latent construction also relies on 3D grid structure.
GSD \cite{MuZGWLWXDYC2024} uses a Gaussian diffusion prior with view guidance at sampling time, but does not train a dedicated multi-view conditional model for sparse-view reconstruction.
DiffSplat \cite{LinPYLM2025} and GaussianObject \cite{YangLFLXZST2024} leverage 2D diffusion priors for Gaussian generation and refinement.
While generative methods can produce sharp and high-quality completions in unseen regions, they are prone to ignoring or modulating the conditioning signals.

\inlineheading{Concurrent Work} 
ReconViaGen \cite{ChangYWCZLLZH2026} conditions a pretrained TRELLIS generator with VGGT features in a voxel latent space.
Unlike ReconViaGen, our approach operates \emph{directly} on Gaussian parameters, without an intermediate voxel representation or multiple large pretrained models.

\section{Free-Range Gaussians}%
\label{sec:method}

\begin{figure}[t]
    \centering
    \includegraphics[trim=0cm 12.0cm 15.5cm 0cm, clip=true, width=\linewidth]{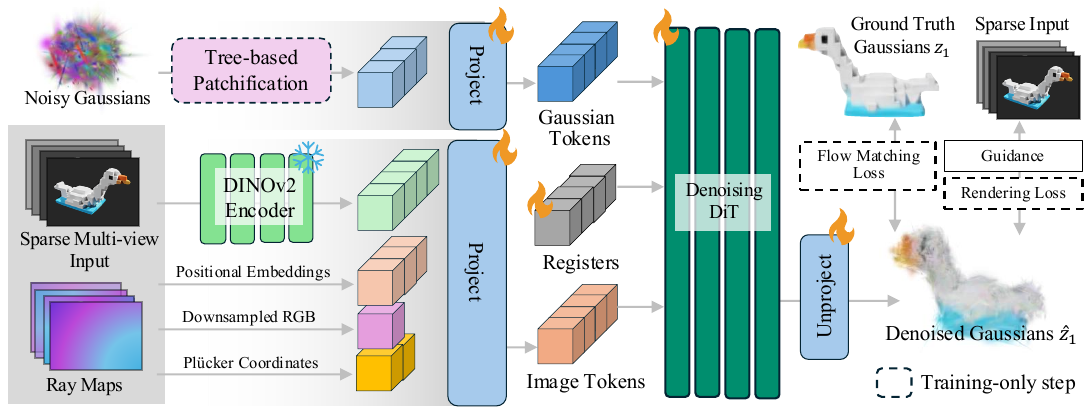}
    \caption{\label{fig:pipeline}%
        \textbf{Method overview.}
        We pose reconstruction as a generative process based on flow matching.
        Conditioned on sparse image features, a diffusion transformer denoises a set of non-grid-aligned 3D Gaussians from random noise (\cref{sec:gen_recon}).
        A level-of-detail tree enables Gaussian patchification for tractable sequence lengths (\cref{sec:hier_patch}).
        A timestep-weighted rendering loss, photometric gradient guidance, and classifier-free guidance steer the output toward consistent reconstructions (\cref{sec:recon_guided}).
    }
    \vspace{-1.5em}
\end{figure}

Our goal is to train a feed-forward model for non-grid-aligned 3D Gaussian reconstruction.
Given a sparse set $\mathcal{I}$ of posed multi-view images of an object, we aim to predict a set $\mathcal{G}$ of 3D Gaussians that faithfully reconstructs the object.
We do this without imposing any pixel- or voxel-based structure on the Gaussians by framing reconstruction as a generative process based on flow matching (\cref{sec:gen_recon}).
Starting from random noise, the model iteratively refines a set of Gaussian parameters conditioned on the input views~$\mathcal{I}$.
This generative formulation has two advantages.
First, it allows the model to produce more plausible completions in unobserved regions 
than is possible with a regression loss.
Second, it enables direct supervision using data that is not anchored to a voxel or pixel grid.
To make the generative process computationally tractable, we organize the Gaussians into a hierarchical tree structure that reduces the number of tokens fed to the denoising model (\cref{sec:hier_patch}).
Finally, we introduce guidance mechanisms to steer generation toward outputs that are consistent with the input views (\cref{sec:recon_guided}).
\Cref{fig:pipeline} provides an overview of our pipeline.

\begin{figure}[t]
\centering\includegraphics[trim=0cm 10.0cm 14.5cm 0.0cm, clip=true, width=1.0\linewidth]{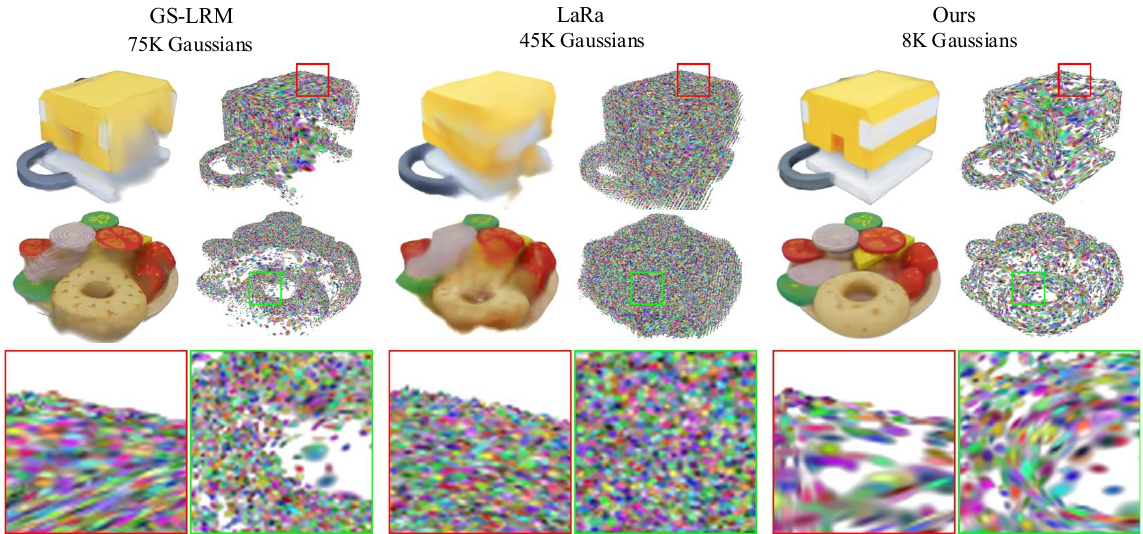}
\caption{\textbf{Generative reconstruction is efficient and complete.}
We visualize reconstruction density by rendering scaled-down Gaussians with random colors.
Pixel-aligned methods like GS-LRM \cite{ZhangBTXZSX2024} leave holes in unobserved regions despite a highly redundant representation.
Voxel-aligned methods like LaRa \cite{ChenXETG2024} are more complete but predict blurry conditional means for unseen parts.
Our generative approach produces sharp, plausible completions with far fewer non-grid-aligned Gaussians.}%
\label{fig:non-pixel-aligned}
\vspace{-2em}
\end{figure}

\vspace{-1em}
\subsection{3D Gaussians via Generative Reconstruction}%
\label{sec:gen_recon}

We frame sparse-view reconstruction as a generative task to address two shortcomings of existing work.
First, regression-based methods trained with L2 losses \cite{ChenXZZXYS2021, WangWGSZBMSF2021, YuYTK2021, ChenXETG2024, ZiwenTZBLHFX2025} predict the conditional expectation $\mathbb{E}[\mathcal{G}\mid\mathcal{I}]$, which averages over all plausible completions and produces blurry results in unseen regions (\cref{fig:non-pixel-aligned}).
A generative model instead samples from the conditional distribution $P(\mathcal{G}\mid\mathcal{I})$, enabling sharp completions.
Second, existing feed-forward methods anchor each Gaussian to either a voxel \cite{ChangYWCZLLZH2026, ChenXETG2024, WangCZLWFQZCZ2025} or an input pixel \cite{SzymaRV2024, TangCCWZL2024, ZhangBTXZSX2024, ChenXZZPGCC2024, CharaLTS2024}, producing redundant grid-aligned outputs.
We instead supervise directly with ground-truth 3D Gaussians, freeing the model from any grid structure.

However, a fundamental challenge in directly regressing Gaussian parameters with a feed-forward model is establishing \emph{correspondences} between predicted and ground-truth Gaussians.
One solution is permutation-invariant losses (\eg, using Chamfer or Earth Mover's distance), but these are imperfect proxies for reconstruction quality, and they become computationally expensive for large sets of Gaussians.
Inspired by Nichol \etal \cite{NichoJDMC2022}, who generate point clouds via diffusion over point coordinates, we sidestep this issue by framing reconstruction as conditional generation via flow matching \cite{LipmaCBNL2023, LiuGl2023}.
Correspondence is established \emph{by construction}: each noisy training sample interpolates between a specific ground-truth Gaussian and noise, and the model learns to recover the ground truth with the ordering kept \emph{fixed}.
In summary, a generative formulation enables detailed and plausible reconstruction in unseen regions and avoids the need to compute expensive permutation-invariant losses to supervise non-grid-aligned Gaussians.

Describing the process in detail, we define a linear probability path between a standard normal prior $\mathcal{N}(0, \mathbf{I})$ and the data distribution of 3D Gaussians, where $\mathbf{I}$ is the identity covariance matrix.
Then, given a ground-truth 3D Gaussian $\mathbf{z}_1$ and noise $\boldsymbol\epsilon \sim \mathcal{N}(0, \mathbf{I})$, the interpolated sample at time $t \in [0, 1]$ is $\mathbf{z}_t = (1 - t) \, \boldsymbol\epsilon + t \, \mathbf{z}_1$.
The model $f_{\boldsymbol\theta}$ predicts the clean sample directly ($x$-prediction \cite{li2025backtobasics}), yielding $\hat{\mathbf{z}}_1 = f_{\boldsymbol\theta}(\mathbf{z}_t, t, \mathcal{I})$, from which the velocity field $\mathbf{v}_t = (\hat{\mathbf{z}}_1 - \mathbf{z}_t) / (1 - t)$ is derived.
At inference, starting from pure noise, this velocity is integrated via Euler steps $\mathbf{z}_{t+\Delta t} = \mathbf{z}_t + \Delta t \cdot \mathbf{v}_t$,
to obtain the final prediction.
We use the standard parameterization of 3D Gaussians by mean position, log-scale, quaternion rotation, logit-opacity, and RGB color.
We normalize each parameter independently using per-parameter mean and standard deviation computed over the training set.

\subsection{Hierarchical Gaussians for Efficient Training}%
\label{sec:hier_patch}

High-quality Gaussian reconstructions typically require more than 50K Gaussians per object, making direct flow matching over the full set prohibitive due to the quadratic cost of self-attention.
We address this by organizing the Gaussians into a tree-based hierarchy, where each node represents a Gaussian and each level of the tree yields a progressively finer representation of the object.
This allows us to train the model at a level that balances detail with tractability.
Furthermore, as the tree structure provides a notion of spatial proximity, it enables us to patchify the sequence \cite{ChenGZSL2025, kim2026ddit} by grouping sibling nodes into a single token.

\begin{figure}[t]
\centering
\includegraphics[trim=.9cm 10.7cm 8.7cm 0cm, clip=true, width=1.0\linewidth]{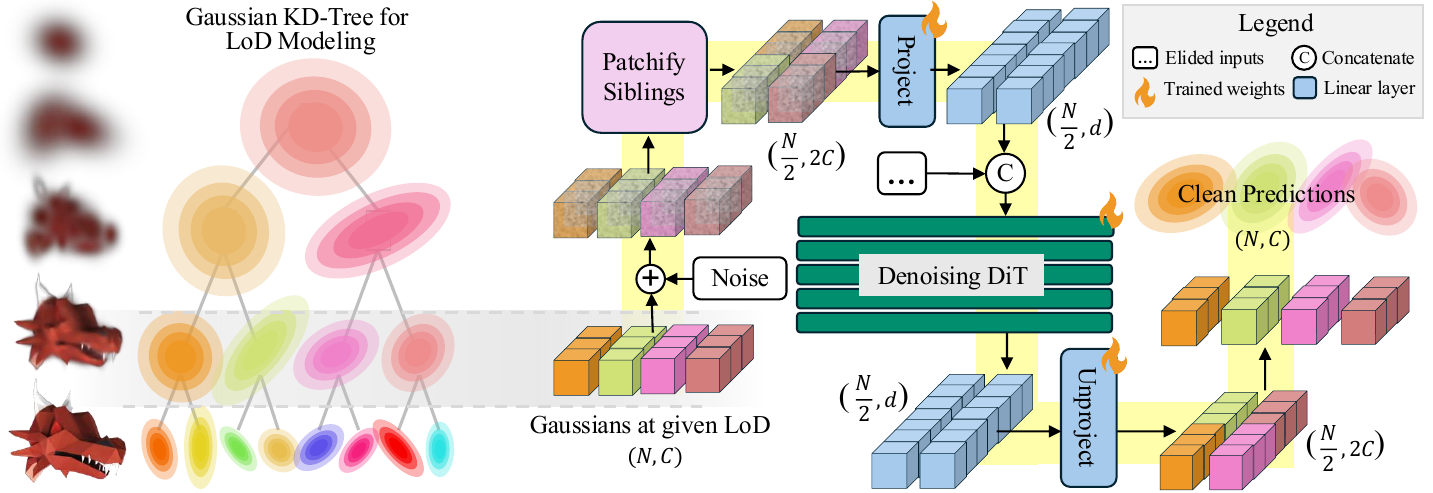}
\caption{
\textbf{Patchification with hierarchical Gaussians.}
A binary tree over the ground-truth Gaussians provides level-of-detail slices at each depth, allowing us to keep the Gaussian count low during training while preserving a faithful object representation.
At the chosen depth, pairs of sibling nodes are merged into single transformer tokens, halving the sequence length while preserving spatial locality.}
\label{fig:patching}
\vspace{-1em}
\end{figure}

Following Kerbl \etal \cite{KerblMKWLD2024}, we construct a tree-based hierarchy over the ground-truth Gaussians of each training object.
Each parent node is a weighted combination of its two children, producing a coarser Gaussian; depth $l$ thus yields $2^l$ Gaussians at a uniform level of detail (\cref{fig:patching}).
Only the Gaussians at a chosen depth serve as model input; parent nodes merely define the hierarchy.
This enables a coarse-to-fine curriculum.
Our Diffusion Transformer (DiT) operates on Gaussian tokens without positional encodings, and is agnostic to sequence length and order, so a trained model can be fine-tuned on different numbers of Gaussians.
Concretely, we first train at depth 11 (2K Gaussians, 224$\times$224 images), then fine-tune at depth 13 (8K Gaussians, 512$\times$512 images), progressively increasing both representational resolution and image-supervision resolution.

To further reduce the sequence length, we exploit the tree structure to patchify the input Gaussians.
Patchification groups neighboring elements into a single token and is standard in pixel-space generative models \cite{ChenGZSL2025, kim2026ddit}. But it does not directly apply to non-grid-aligned Gaussians, which lack a regular spatial structure.
The tree provides a natural substitute: we group pairs of sibling Gaussians into single tokens, \emph{halving} the sequence from $\mathbb{R}^{N \times C}$ to $\mathbb{R}^{(N/2) \times 2C}$, where $N$ is the number of Gaussians at the chosen depth and $C{=}14$ is the per-Gaussian parameter dimensionality.
Since siblings are spatially proximate, they form a coherent group and provide the transformer with a natural structural prior.
A linear layer then maps the patched vectors to the transformer hidden dimension $d \!=\! 1024$.
At the output, the process is reversed.
Thus, we use the hierarchy for (1)~selecting LoD slices for coarse-to-fine training, and (2)~defining pairs of tokens for patchification.

\subsection{Reconstruction-Guided Flow Matching}%
\label{sec:recon_guided}

While improving training efficiency, the LoD slicing described in the previous section provides a coarser signal than the original reconstruction. We address this via three complementary mechanisms to recover visual details in the generated output:
(1)~a timestep-weighted photometric loss that supervises rendered images once predictions become reliable,
(2)~photometric gradient guidance that steers Gaussian parameters toward lower rendering error at each denoising step, and
(3)~classifier-free guidance \cite{HoS2021} to amplify the influence of the input views.

\inlineheading{Timestep-Weighted Multi-View Rendering Loss}
We add photometric supervision to sharpen details beyond what the flow matching loss alone provides. Thus, at each denoising step, we compute the L1 error between renderings of the predicted $\hat{\mathbf{z}}_1$ and both input and ground-truth images.
However, when $t$ is close to zero, $\hat{\mathbf{z}}_1$ is unreliable and a photometric loss provides noisy gradients.
Therefore, we weight the rendering losses $\mathcal{R}_{\text{Held}}$ and $\mathcal{R}_{\text{Seen}}$, over held-out ground-truth and observed input views, respectively, with a monotonically increasing function
\[
w(t) = 50 \cdot \left(\min\left(1, \frac{t}{0.9}\right)\right)^5
\]
that concentrates supervision on later stages, when the model's prediction is close to $\mathbf{z}_1$.
The combined training loss is:
\begin{equation*}
\label{eq:fm-loss}
\mathcal{L} = \mathcal{L}_\text{FM} + w(t) \cdot \big( \mathcal{R}_{\text{Held}} + \mathcal{R}_{\text{Seen}}\big),\ \text{where}\
\mathcal{L}_\text{FM} = \Big\lVert \left( \frac{\hat{\mathbf{z}}_1 - \mathbf{z}_t}{1-t}\right) - (\mathbf{z}_1 - \boldsymbol\epsilon) \Big\rVert_2^2.
\end{equation*}

\inlineheading{Photometric Gradient Guidance}
Inspired by Mu \etal \cite{MuZGWLWXDYC2024}, we steer the model toward predictions that better match the observed input views through explicit reconstruction feedback.
At each denoising step, we compute $\nabla_{\mathbf{z}} \mathcal{R}_{\text{Seen}}$, the gradient of the photometric loss with respect to the Gaussian parameters.
These per-Gaussian gradients are then concatenated with the noisy inputs as additional channels.
This provides the model with a direct signal about how each Gaussian should be adjusted.
This is done during both training and inference.
$\quad$
During inference, we additionally incorporate these gradients into the Euler update with a guidance scale $\lambda_\text{PG}$.
This steers the denoising trajectory toward reconstructions that match the input views:
$
\mathbf{z}_{t + \Delta t} = \mathbf{z}_t + \Delta t \cdot \mathbf{v}_t - \lambda_\text{PG} \cdot \nabla_{\mathbf{z}} \mathcal{R}_\text{Seen}.
$ For our experiments, we use $\lambda_\textrm{PG} \!=\! 50$
(see \cref{table:ablations} for an analysis).

\inlineheading{Classifier-Free Guidance}
We use classifier-free guidance \cite{HoS2021} to further amplify the influence of the conditioning views.
Specifically, we train the model to produce unconditional output by zeroing out the image features with probability 10\% during training.
Then, at inference time, the guided prediction is:
$
\hat{\mathbf{z}}_1^\text{guided} = \hat{\mathbf{z}}_1^\text{uncond} + s \cdot (\hat{\mathbf{z}}_1^\text{cond} - \hat{\mathbf{z}}_1^\text{uncond})
$, where $s > 1$ is the guidance scale.

\subsection{Model Architecture}
\label{sec:model_architecture}

Our model is based on the DiT architecture \cite{PeeblX2023}.
The input images $\mathcal{I}$ are processed by a frozen DINOv2-Base \cite{OquabDMVSKFHMEABGHHLMRSSXJMLJB2024} encoder, producing patch tokens of dimension 768.
These are concatenated along the channel dimension with positional embeddings (32 channels), downsampled RGB values (3 channels), and Plücker ray coordinates (6 channels) of each view.
The image tokens are projected to the hidden dimension of 1024 via a linear layer, and concatenated with the patchified noisy Gaussian tokens and 32 learnable embedding tokens acting as registers \cite{darcet2024vision}.
The combined sequence is processed by 18 transformer blocks.
Each block consists of self-attention with 16 64-dimensional heads, AdaLN \cite{PeeblX2023} conditioning on the timestep $t$, and a GEGLU feed-forward network.
The transformer output is mapped to the input size via a linear projection.
QK-normalization \cite{henry2020query} via RMSNorm \cite{zhang2019root} is applied for training stability.

\ifdefined\gs
\else
  \def\gsreconviagen{500}
  \def\gslgm{65}
  \def\gslara{45}
  \def\gsgslrm{70}
  \def\gsours{8}

  \fpmin{\gsmin}{\gsreconviagen}{\gslgm}{\gslara}{\gsgslrm}{\gsours}
  \fpmax{\gsmax}{\gsreconviagen}{\gslgm}{\gslara}{\gsgslrm}{\gsours}
  \fpnorminvlog{\gs}{\gsmin}{\gsmax}
\fi

\def\psnrpobjreconviagen{21.33}
\def\psnrpobjlgm{22.88}
\def\psnrpobjlara{27.79}
\def\psnrpobjgslrm{27.90}
\def\psnrpobjours{29.92}

\def\psnrpgsoreconviagen{17.67}
\def\psnrpgsolgm{19.65}
\def\psnrpgsolara{25.64}
\def\psnrpgsogslrm{27.57}
\def\psnrpgsoours{28.08}

\fpmin{\psnrpobjmin}{\psnrpobjreconviagen}{\psnrpobjlgm}{\psnrpobjlara}{\psnrpobjgslrm}{\psnrpobjours}

\fpmax{\psnrpobjmax}{\psnrpobjreconviagen}{\psnrpobjlgm}{\psnrpobjlara}{\psnrpobjgslrm}{\psnrpobjours}

\fpnorm{\psnrpobj}{\psnrpobjmin}{\psnrpobjmax}

\fpmin{\psnrpgsomin}{\psnrpgsoreconviagen}{\psnrpgsolgm}{\psnrpgsolara}{\psnrpgsogslrm}{\psnrpgsoours}

\fpmax{\psnrpgsomax}{\psnrpgsoreconviagen}{\psnrpgsolgm}{\psnrpgsolara}{\psnrpgsogslrm}{\psnrpgsoours}

\fpnorm{\psnrpgso}{\psnrpgsomin}{\psnrpgsomax}

\def\dsobjreconviagen{18.21}
\def\dsobjlgm{17.05}
\def\dsobjlara{12.39}
\def\dsobjgslrm{10.33}
\def\dsobjours{9.24}

\def\dsgsoreconviagen{24.35}
\def\dsgsolgm{26.41}
\def\dsgsolara{22.28}
\def\dsgsogslrm{16.11}
\def\dsgsoours{15.50}

\fpmin{\dsobjmin}{\dsobjreconviagen}{\dsobjlgm}{\dsobjlara}{\dsobjgslrm}{\dsobjours}

\fpmax{\dsobjmax}{\dsobjreconviagen}{\dsobjlgm}{\dsobjlara}{\dsobjgslrm}{\dsobjours}

\fpnorminv{\dsobj}{\dsobjmin}{\dsobjmax}

\fpmin{\dsgsomin}{\dsgsoreconviagen}{\dsgsolgm}{\dsgsolara}{\dsgsogslrm}{\dsgsoours}

\fpmax{\dsgsomax}{\dsgsoreconviagen}{\dsgsolgm}{\dsgsolara}{\dsgsogslrm}{\dsgsoours}

\fpnorminv{\dsgso}{\dsgsomin}{\dsgsomax}

\def\dinoobjreconviagen{68.71}
\def\dinoobjlgm{66.46}
\def\dinoobjlara{71.72}
\def\dinoobjgslrm{77.57}
\def\dinoobjours{81.80}

\def\dinogsoreconviagen{68.04}
\def\dinogsolgm{58.01}
\def\dinogsolara{60.20}
\def\dinogsogslrm{72.20}
\def\dinogsoours{72.24}

\fpmin{\dinoobjmin}{\dinoobjreconviagen}{\dinoobjlgm}{\dinoobjlara}{\dinoobjgslrm}{\dinoobjours}

\fpmax{\dinoobjmax}{\dinoobjreconviagen}{\dinoobjlgm}{\dinoobjlara}{\dinoobjgslrm}{\dinoobjours}

\fpnorm{\dinoobj}{\dinoobjmin}{\dinoobjmax}

\fpmin{\dinogsomin}{\dinogsoreconviagen}{\dinogsolgm}{\dinogsolara}{\dinogsogslrm}{\dinogsoours}

\fpmax{\dinogsomax}{\dinogsoreconviagen}{\dinogsolgm}{\dinogsolara}{\dinogsogslrm}{\dinogsoours}

\fpnorm{\dinogso}{\dinogsomin}{\dinogsomax}

\def\fidobjreconviagen{66.07}
\def\fidobjlgm{88.43}
\def\fidobjlara{93.19}
\def\fidobjgslrm{65.06}
\def\fidobjours{43.58}

\def\fidgsoreconviagen{23.32}
\def\fidgsolgm{54.96}
\def\fidgsolara{69.18}
\def\fidgsogslrm{34.13}
\def\fidgsoours{27.69}

\fpmin{\fidobjmin}{\fidobjreconviagen}{\fidobjlgm}{\fidobjlara}{\fidobjgslrm}{\fidobjours}

\fpmax{\fidobjmax}{\fidobjreconviagen}{\fidobjlgm}{\fidobjlara}{\fidobjgslrm}{\fidobjours}

\fpnorminv{\fidobj}{\fidobjmin}{\fidobjmax}

\fpmin{\fidgsomin}{\fidgsoreconviagen}{\fidgsolgm}{\fidgsolara}{\fidgsogslrm}{\fidgsoours}

\fpmax{\fidgsomax}{\fidgsoreconviagen}{\fidgsolgm}{\fidgsolara}{\fidgsogslrm}{\fidgsoours}

\fpnorminv{\fidgso}{\fidgsomin}{\fidgsomax}

\begin{table*}[tb]
\caption{\textbf{Quantitative comparison of partially observed objects.}
We evaluate the case when four input views are concentrated in a hemisphere, leaving parts of the object unobserved.
The values for DreamSim (DS) \cite{FuTSCZDI2023} and DINO similarity are $\times 100$.
Our method shows strong performance on all metrics, with an order-of-magnitude fewer Gaussians (\#GS).
We highlight the metrics in blue, proportional to their percentile.}%
\label{table:results-partial}
\centering
\begin{threeparttable}
\renewcommand{\arraystretch}{1.15}
\resizebox{\linewidth}{!}{
\begin{tabular}{l@{\hspace{6pt}}r@{\hspace{6pt}}cccc@{\hspace{6pt}}cccc}
\toprule
& & \multicolumn{4}{c@{\hspace{6pt}}}{Objaverse} & \multicolumn{4}{c}{GSO} \\
\cmidrule(lr){3-6} \cmidrule(lr){7-10}
Method & \#GS & PSNR\up &  DINO\up & DS\down & FID\down & PSNR\up &  DINO\up & DS\down & FID\down \\
\midrule
\tnote{$\star$} ReconViaGen & \gs{\gsreconviagen}K & \hspace{-2mm}\psnrpobj{\psnrpobjreconviagen} &
\hspace{-2mm}\dinoobj{\dinoobjreconviagen} &
\hspace{-2mm}\dsobj{\dsobjreconviagen} &
\hspace{-2mm}\fidobj{\fidobjreconviagen} &
\hspace{-2mm}\psnrpgso{\psnrpgsoreconviagen} &
\hspace{-2mm}\dinogso{\dinogsoreconviagen} &
\hspace{-2mm}\dsgso{\dsgsoreconviagen} &
\hspace{-2mm}\textbf{\fidgso{\fidgsoreconviagen}} \\
LGM & \gs{\gslgm}K & \hspace{-2mm}\psnrpobj{\psnrpobjlgm} &
\hspace{-2mm}\dinoobj{\dinoobjlgm} &
\hspace{-2mm}\dsobj{\dsobjlgm} &
\hspace{-2mm}\fidobj{\fidobjlgm} &
\hspace{-2mm}\psnrpgso{\psnrpgsolgm} &
\hspace{-2mm}\dinogso{\dinogsolgm} &
\hspace{-2mm}\dsgso{\dsgsolgm}&
\hspace{-2mm}\fidgso{\fidgsolgm} \\
LaRa & \gs{\gslara}K & \hspace{-2mm}\psnrpobj{\psnrpobjlara} &
\hspace{-2mm}\dinoobj{\dinoobjlara} &
\hspace{-2mm}\dsobj{\dsobjlara} &
\hspace{-2mm}\fidobj{\fidobjlara} &
\hspace{-2mm}\psnrpgso{\psnrpgsolara} &
\hspace{-2mm}\dinogso{\dinogsolara} &
\hspace{-2mm}\dsgso{\dsgsolara} &
\hspace{-2mm}\fidgso{\fidgsolara} \\
GS-LRM & \gs{\gsgslrm}K & \hspace{-2mm}\psnrpobj{\psnrpobjgslrm} &
\hspace{-2mm}\dinoobj{\dinoobjgslrm} &
\hspace{-2mm}\dsobj{\dsobjgslrm} &
\hspace{-2mm}\fidobj{\fidobjgslrm} &
\hspace{-2mm}\psnrpgso{\psnrpgsogslrm} &
\hspace{-2mm}\dinogso{\dinogsogslrm} &
\hspace{-2mm}\dsgso{\dsgsogslrm} &
\hspace{-2mm}\fidgso{\fidgsogslrm} \\
Ours & \textbf{\gs{\gsours}K} & \hspace{-2mm}\textbf{\psnrpobj{\psnrpobjours}} &
\hspace{-2mm}\textbf{\dinoobj{\dinoobjours}} &
\hspace{-2mm}\textbf{\dsobj{\dsobjours}} &
\hspace{-2mm}\textbf{\fidobj{\fidobjours}} & \hspace{-2mm}\textbf{\psnrpgso{\psnrpgsoours}} &
\hspace{-2mm}\textbf{\dinogso{\dinogsoours}} &
\hspace{-2mm}\textbf{\dsgso{\dsgsoours}} &
\hspace{-2mm}\fidgso{\fidgsoours} \\
\bottomrule
\end{tabular}}
\begin{tablenotes}
            \item[] \hfill \tnote{$\star$} \scriptsize{Concurrent work.} 
        \end{tablenotes}
\end{threeparttable}
\vspace{-1.5em}
\end{table*}

\vspace{-0.5em}
\subsection{Training Strategy}%
\label{sec:training}

We train on 3D Gaussian reconstructions of ${\sim}$140K Objaverse objects \cite{DeitkSSWMVSEKF2023}.
Timesteps $t$ follow a cosine schedule \cite{nichol2021improved}.
As described in \cref{sec:hier_patch}, we use a coarse-to-fine curriculum: first training with 2K Gaussians at 224$\times$224 resolution, then fine-tuning with 8K Gaussians at 512$\times$512.
We use AdamW with a learning rate of $10^{-4}$, cosine decay to $5 \times 10^{-6}$, and 2,000 warmup steps, training for 700K iterations on 64 NVIDIA H200 GPUs for 6 days.

\subsection{Inference}%
\label{sec:arch_inference}

Starting from $\mathbf{z}_{t{=}0} = \boldsymbol\epsilon \sim \mathcal{N}(0, \mathbf{I})$, we integrate the learned velocity field with 50 Euler steps ($\Delta t{=}0.02$).
At each step, the model predicts $\hat{\mathbf{z}}_1$, from which the velocity is computed and the state updated.
Classifier-free guidance and photometric gradient guidance (\cref{sec:recon_guided}) are applied at every step.
The final $\hat{\mathbf{z}}_1$ is denormalized to obtain the predicted Gaussian parameters $\hat{\mathcal{G}}$, which can be directly rendered via the 3DGS rasterizer.
Single object inference takes approximately 26 seconds on a single H200 GPU.

\section{Experiments}%
\label{sec:experiments}

\label{sec:datasets}

\inlineheading{Training Data}
Following ShapeSplat \cite{MaLRSKGGP2025}, we created a dataset of ${\sim}$140K objects from Objaverse.
Each object is rendered from 40 views at 512$\times$512, and then reconstructed using 3DGS-MCMC optimization \cite{kheradmand2024mcmc}.
A hierarchical LoD tree is precomputed for each object using Kerbl~\etal's method \cite{KerblMKWLD2024}.

\inlineheading{Test Data}
We evaluate on two benchmarks:
(1)~Objaverse \cite{DeitkSSWMVSEKF2023} (500 held-out objects) processed as described above, and
(2)~Google Scanned Objects (GSO) \cite{DownsFKKHRMV2022} (1,030 objects), a dataset of real scanned household items providing an out-of-distribution test of generalization.

\begin{wrapfigure}{r}{0.4\textwidth}
    \vspace{-28pt}
    \centering
    \includegraphics[width=\linewidth]{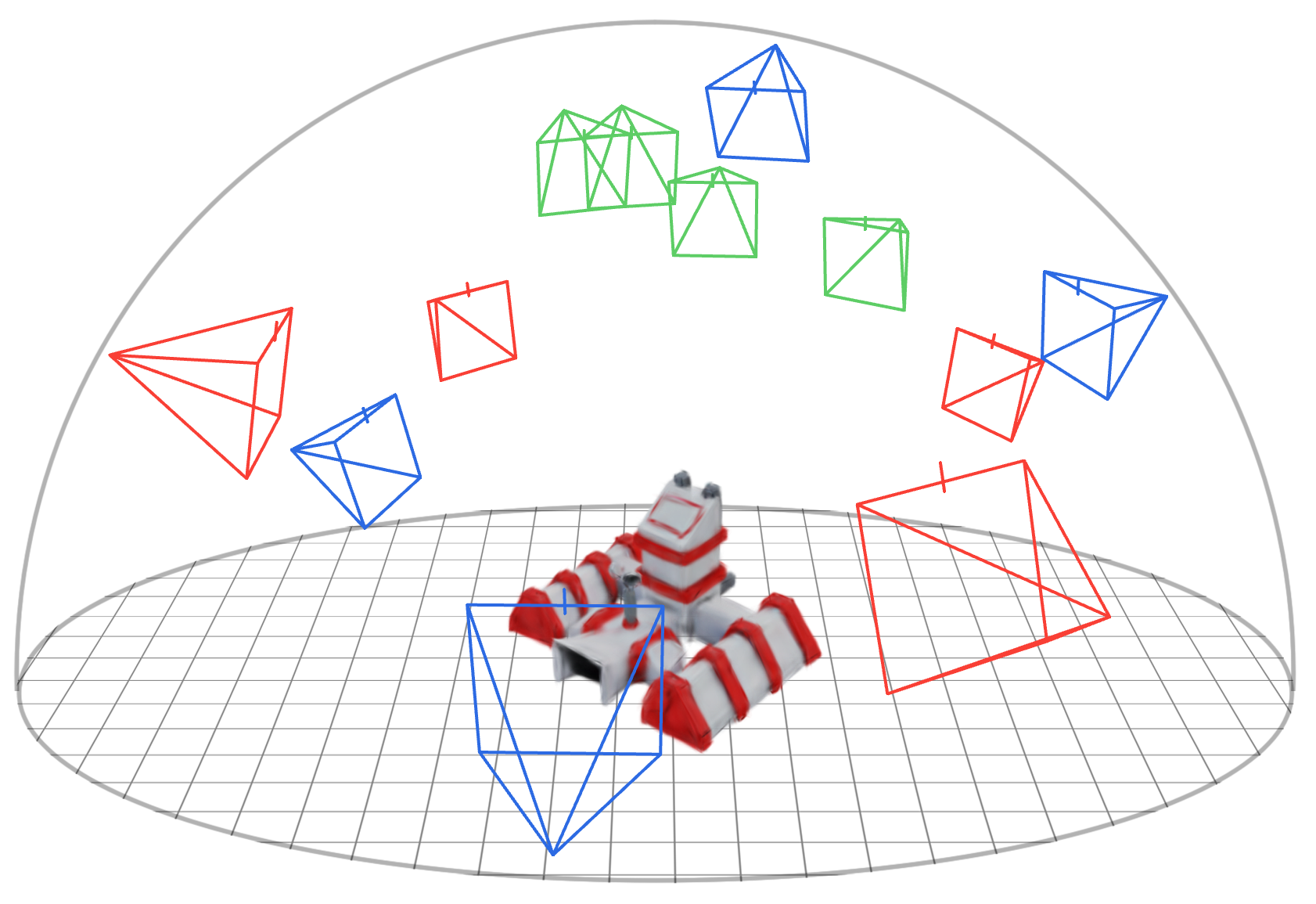}
    \vspace{-10pt}
    \caption{\textcolor{blue}{Full observation}, \textcolor{green!70!black}{Partial observation}, and \textcolor{red}{Validation} cameras.}
    \vspace{-20pt}
    \label{fig:camera-config}
\end{wrapfigure}
\inlineheading{Input View Configurations}
We evaluate two input-view settings (\cref{fig:camera-config}). \textbf{\textcolor{blue}{Full observation:}}
following LaRa \cite{ChenXETG2024}, we use K-means to cluster the cameras into four clusters and sample one view per cluster.
This ensures sufficient angular coverage of the input views. \textbf{\textcolor{green!70!black}{Partial observation:}}
all four views are drawn from a single cluster, concentrating observations in one region and leaving other parts of the object unobserved.
This setting is specifically designed to test how methods handle unobserved regions.
\textbf{\textcolor{red}{Validation views:}} for both input settings, we use the same fixed validation-view set, sampled to provide broad 360$^\circ$ object coverage; this set is similar to the full-observation pattern but not identical.

\ifdefined\gs
\else
  \def\gsreconviagen{500}
  \def\gslgm{65}
  \def\gslara{45}
  \def\gsgslrm{70}
  \def\gsours{8}

  \fpmin{\gsmin}{\gsreconviagen}{\gslgm}{\gslara}{\gsgslrm}{\gsours}
  \fpmax{\gsmax}{\gsreconviagen}{\gslgm}{\gslara}{\gsgslrm}{\gsours}
  \fpnorminvlog{\gs}{\gsmin}{\gsmax}
\fi

\def\psnrobjreconviagen{22.54}
\def\psnrobjlgm{26.65}
\def\psnrobjlara{31.91}
\def\psnrobjgslrm{31.03}
\def\psnrobjours{31.66}

\def\psnrgsoreconviagen{18.29}
\def\psnrgsolgm{23.64}
\def\psnrgsolara{29.15}
\def\psnrgsogslrm{31.13}
\def\psnrgsoours{31.49}

\fpmin{\psnrobjmin}{\psnrobjreconviagen}{\psnrobjlgm}{\psnrobjlara}{\psnrobjgslrm}{\psnrobjours}

\fpmax{\psnrobjmax}{\psnrobjreconviagen}{\psnrobjlgm}{\psnrobjlara}{\psnrobjgslrm}{\psnrobjours}

\fpnorm{\psnrobj}{\psnrobjmin}{\psnrobjmax}

\fpmin{\psnrgsomin}{\psnrgsoreconviagen}{\psnrgsolgm}{\psnrgsolara}{\psnrgsogslrm}{\psnrgsoours}

\fpmax{\psnrgsomax}{\psnrgsoreconviagen}{\psnrgsolgm}{\psnrgsolara}{\psnrgsogslrm}{\psnrgsoours}

\fpnorm{\psnrgso}{\psnrgsomin}{\psnrgsomax}

\def\ssimobjreconviagen{94.01}
\def\ssimobjlgm{95.50}
\def\ssimobjlara{97.12}
\def\ssimobjgslrm{97.10}
\def\ssimobjours{97.14}

\def\ssimgsoreconviagen{90.42}
\def\ssimgsolgm{92.35}
\def\ssimgsolara{95.60}
\def\ssimgsogslrm{95.00}
\def\ssimgsoours{95.13}

\fpmin{\ssimobjmin}{\ssimobjreconviagen}{\ssimobjlgm}{\ssimobjlara}{\ssimobjgslrm}{\ssimobjours}

\fpmax{\ssimobjmax}{\ssimobjreconviagen}{\ssimobjlgm}{\ssimobjlara}{\ssimobjgslrm}{\ssimobjours}

\fpnorm{\ssimobj}{\ssimobjmin}{\ssimobjmax}

\fpmin{\ssimgsomin}{\ssimgsoreconviagen}{\ssimgsolgm}{\ssimgsolara}{\ssimgsogslrm}{\ssimgsoours}

\fpmax{\ssimgsomax}{\ssimgsoreconviagen}{\ssimgsolgm}{\ssimgsolara}{\ssimgsogslrm}{\ssimgsoours}

\fpnorm{\ssimgso}{\ssimgsomin}{\ssimgsomax}

\def\lpipsobjreconviagen{82.27}
\def\lpipsobjlgm{57.71}
\def\lpipsobjlara{43.60}
\def\lpipsobjgslrm{32.80}
\def\lpipsobjours{53.15}

\def\lpipsgsoreconviagen{102.3}
\def\lpipsgsolgm{73.61}
\def\lpipsgsolara{60.70}
\def\lpipsgsogslrm{30.93}
\def\lpipsgsoours{77.25}

\fpmin{\lpipsobjmin}{\lpipsobjreconviagen}{\lpipsobjlgm}{\lpipsobjlara}{\lpipsobjgslrm}{\lpipsobjours}

\fpmax{\lpipsobjmax}{\lpipsobjreconviagen}{\lpipsobjlgm}{\lpipsobjlara}{\lpipsobjgslrm}{\lpipsobjours}

\fpnorminv{\lpipsobj}{\lpipsobjmin}{\lpipsobjmax}

\fpmin{\lpipsgsomin}{\lpipsgsoreconviagen}{\lpipsgsolgm}{\lpipsgsolara}{\lpipsgsogslrm}{\lpipsgsoours}

\fpmax{\lpipsgsomax}{\lpipsgsoreconviagen}{\lpipsgsolgm}{\lpipsgsolara}{\lpipsgsogslrm}{\lpipsgsoours}

\fpnorminv{\lpipsgso}{\lpipsgsomin}{\lpipsgsomax}

\begin{table*}[tb]
\caption{\textbf{Quantitative comparison of fully observed objects.}
We evaluate the case in which four input views provide full coverage of the object.
The values for SSIM and LPIPS are $\times 100$ and $\times 1000$, respectively.
Our method achieves competitive quality with grid-aligned methods while generating an order-of-magnitude fewer Gaussians (\#GS).
We highlight the metrics in blue, proportional to their percentile.}%
\label{table:results-full}
\centering
\begin{threeparttable}
\renewcommand{\arraystretch}{1.15}
\begin{tabular}{l@{\hspace{6pt}}r@{\hspace{6pt}}ccc@{\hspace{6pt}}ccc}
\toprule
& & \multicolumn{3}{c@{\hspace{6pt}}}{Objaverse} & \multicolumn{3}{c}{GSO} \\
\cmidrule(lr){3-5} \cmidrule(lr){6-8}
Method & \#GS & PSNR\up & SSIM\up & LPIPS\down & PSNR\up & SSIM\up & LPIPS\down \\
\midrule
\tnote{$\star$} ReconViaGen~\cite{ChangYWCZLLZH2026} & \gs{\gsreconviagen}K &
\hspace{-2mm}\psnrobj{\psnrobjreconviagen} &
\hspace{-2mm}\ssimobj{\ssimobjreconviagen} &
\hspace{-2mm}\lpipsobj{\lpipsobjreconviagen} &
\hspace{-2mm}\psnrgso{\psnrgsoreconviagen} &
\hspace{-2mm}\ssimgso{\ssimgsoreconviagen} &
\hspace{-2mm}\lpipsgso{\lpipsgsoreconviagen} \\
LGM~\cite{TangCCWZL2024} & \gs{\gslgm}K & \hspace{-2mm}\psnrobj{\psnrobjlgm} &
\hspace{-2mm}\ssimobj{\ssimobjlgm} &
\hspace{-2mm}\lpipsobj{\lpipsobjlgm} &
\hspace{-2mm}\psnrgso{\psnrgsolgm} &
\hspace{-2mm}\ssimgso{\ssimgsolgm} &
\hspace{-2mm}\lpipsgso{\lpipsgsolgm} \\
LaRa~\cite{ChenXETG2024} & \gs{\gslara}K & \textbf{\hspace{-2mm}\psnrobj{\psnrobjlara}} & \hspace{-2mm}{\ssimobj{\ssimobjlara}} & \hspace{-2mm}\lpipsobj{\lpipsobjlara} & \hspace{-2mm}\psnrgso{\psnrgsolara} & \textbf{\hspace{-2mm}\ssimgso{\ssimgsolara}} & \hspace{-2mm}\lpipsgso{\lpipsgsolara} \\
GS-LRM~\cite{ZhangBTXZSX2024} & \gs{\gsgslrm}K & \hspace{-2mm}\psnrobj{\psnrobjgslrm} & \hspace{-2mm}{\ssimobj{\ssimobjgslrm}} & \textbf{\hspace{-2mm}\lpipsobj{\lpipsobjgslrm}} & \hspace{-2mm}\psnrgso{\psnrgsogslrm} & \hspace{-2mm}\ssimgso{\ssimgsogslrm} & \textbf{\hspace{-2mm}\lpipsgso{\lpipsgsogslrm}} \\
Ours & \textbf{\gs{\gsours}K} & \hspace{-2mm}\psnrobj{\psnrobjours} & \hspace{-2mm}\textbf{\ssimobj{\ssimobjours}} & \hspace{-2mm}\lpipsobj{\lpipsobjours} & \textbf{\hspace{-2mm}\psnrgso{\psnrgsoours}} & \hspace{-2mm}\ssimgso{\ssimgsoours} & \hspace{-2mm}\lpipsgso{\lpipsgsoours} \\
\bottomrule
\end{tabular}
\begin{tablenotes}
            \item[] \hfill \tnote{$\star$} \scriptsize{Concurrent work. 
            }
        \end{tablenotes}
\end{threeparttable}
\vspace{-1.5em}
\end{table*}

\begin{figure}[t]
\centering    
\includegraphics[trim=0cm 2.2cm 12.2cm 0cm, clip=true, width=1.0\linewidth]{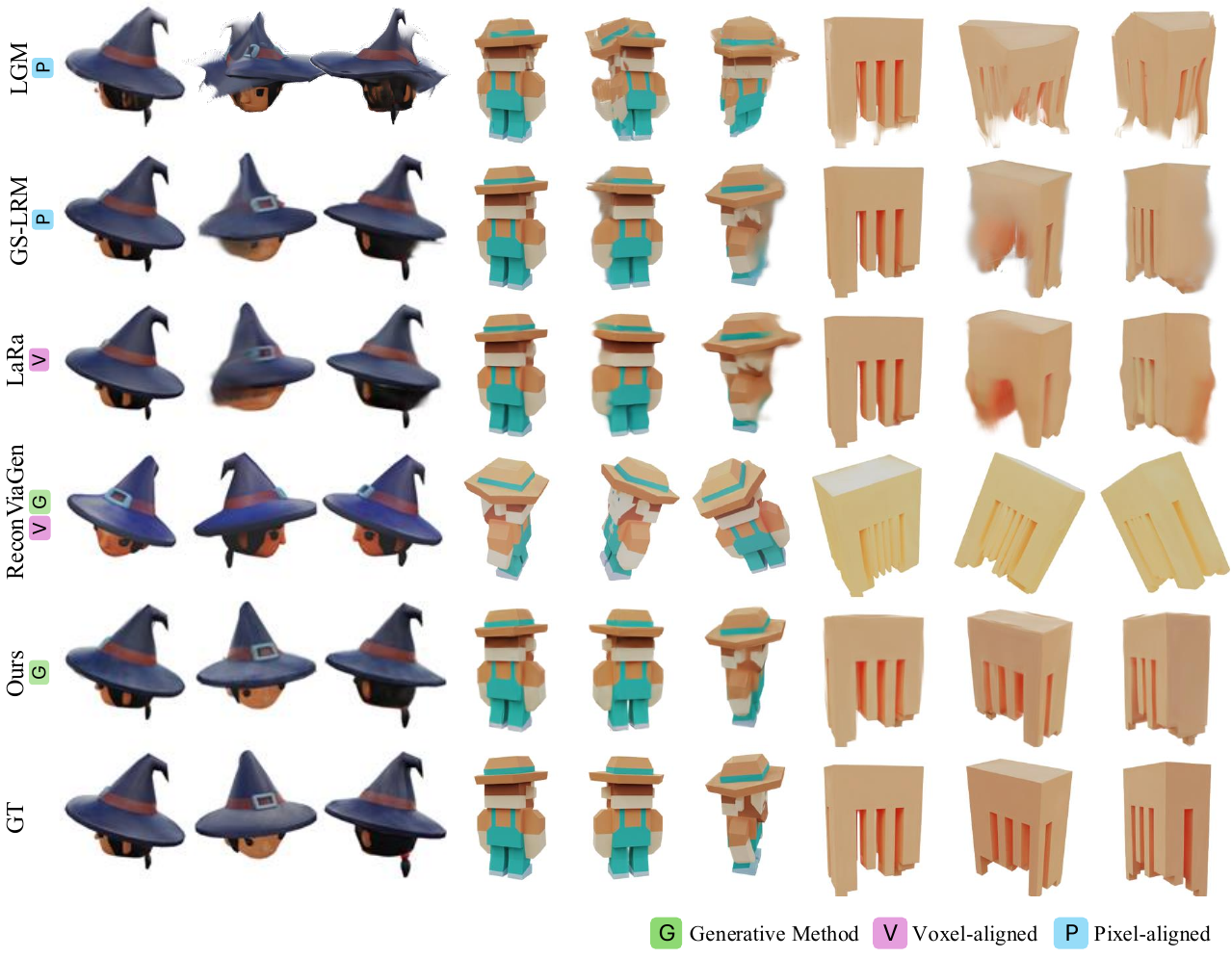}
\caption{
\textbf{Qualitative comparison on the Objaverse dataset \cite{DeitkSSWMVSEKF2023}.}
The pixel-aligned LGM \cite{TangCCWZL2024} and GS-LRM \cite{ZhangBTXZSX2024} produce severe artifacts and holes in unobserved regions,  while LaRa \cite{ChenXETG2024} yields blurry completions.
ReconViaGen's generative reconstruction \cite{ChangYWCZLLZH2026} synthesizes realistic content but is misaligned with the input views.
Our method produces sharp, input-consistent reconstructions for both observed and unobserved regions.}%
\label{fig:results-compare-pixel-aligned}
\end{figure}

\inlineheading{Baselines}
We compare against four feed-forward reconstruction methods spanning the pixel-aligned, volumetric, and generative paradigms:
LGM \cite{TangCCWZL2024}, a U-Net predicting pixel-aligned Gaussians from four input views;
GS-LRM \cite{ZhangBTXZSX2024}, a pixel-aligned transformer with Plücker ray embeddings;
LaRa \cite{ChenXETG2024}, a voxel-based method trained with multi-view rendering loss;
and ReconViaGen \cite{ChangYWCZLLZH2026}, a generative reconstruction method.  As the source code for GS-LRM is not released we use our re-implementation which matches or slightly exceeds the reported results \cite{DownsFKKHRMV2022}).

\begin{figure}[th]
\centering
\includegraphics[trim=0cm 2.2cm 11.0cm 0cm, clip=true, width=1.0\linewidth]{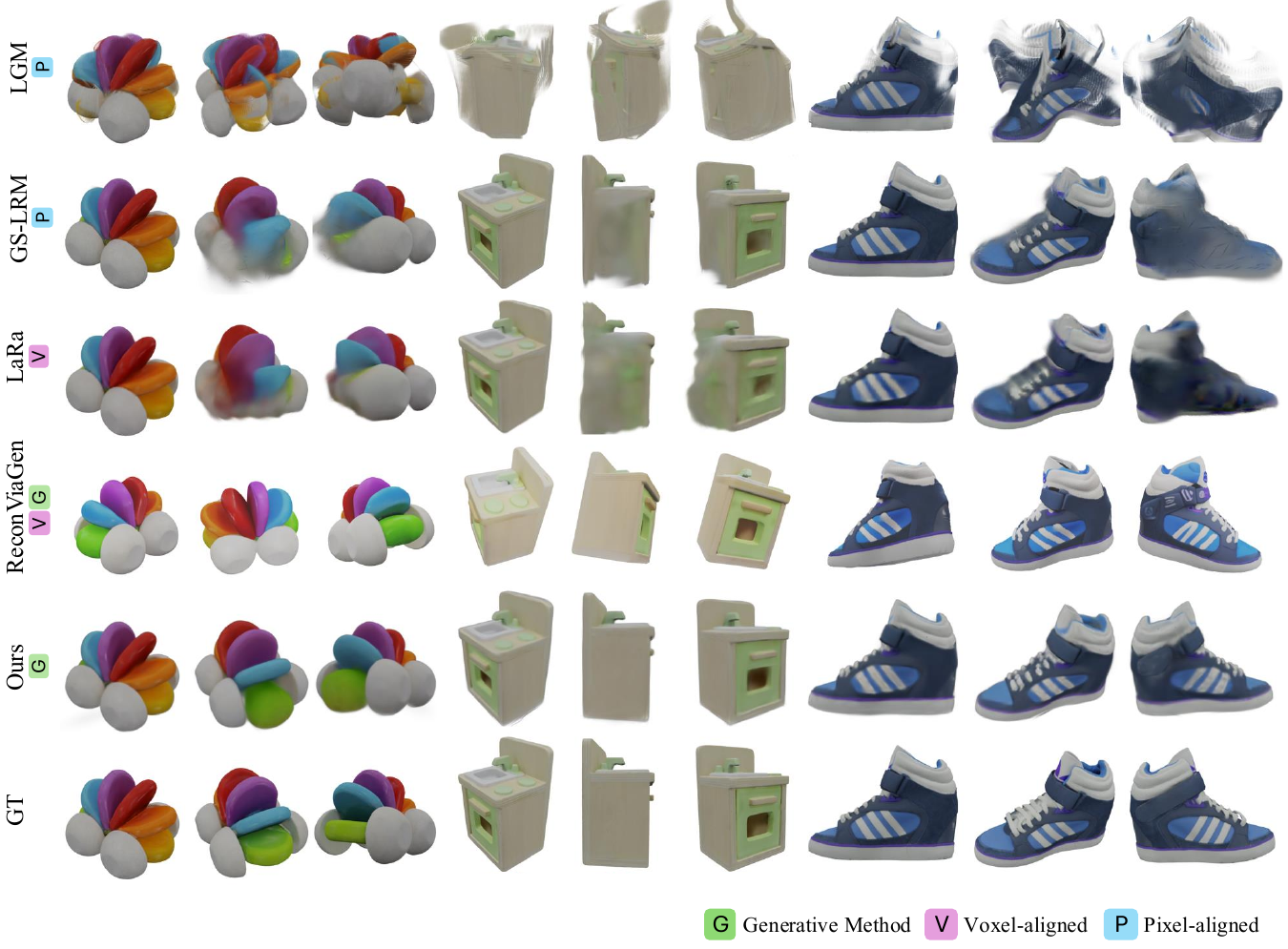}
\caption{\textbf{Qualitative comparison on the GSO dataset.}
Additional examples comparing the input-fidelity and completeness of our method with the baselines.
}%
\label{fig:results-qualitative-b}
\end{figure}

\subsection{Results}

We present quantitative results on both benchmark datasets in
\cref{table:results-partial} and \cref{table:results-full}; qualitative comparisons are shown in \cref{fig:results-compare-pixel-aligned,fig:results-qualitative-b}.
For full observation, we report standard pixel-level metrics (PSNR, SSIM, LPIPS); for partial observation, where unobserved regions dominate, we replace SSIM and LPIPS with perceptual and distributional metrics (DINO, DreamSim, FID) that better capture the quality of hallucinated content.

\inlineheading{Full Observation}
Under full observation (\cref{table:results-full}), our method performs on par with the baselines while using only 8K Gaussians.
Notably, this is an order-of-magnitude fewer Gaussians than all other baselines (45K--500K).

\inlineheading{Partial Observation}
The advantage of our approach is most pronounced under partial observation (\cref{table:results-partial,fig:results-compare-pixel-aligned}), where input views are concentrated on one side, leaving large parts of the object unobserved.
Here, our method outperforms all baselines on most metrics across both datasets, including perceptual (DreamSim \cite{FuTSCZDI2023}, DINOv2-feature \cite{OquabDMVSKFHMEABGHHLMRSSXJMLJB2024}) and distributional (FID \cite{HeuseRUNH2017}) measures.
The strong FID scores reflect a key benefit of the generative formulation: our model produces perceptually realistic completions rather than blurry conditional averages.
Pixel-aligned methods (LGM, GS-LRM) degrade significantly, as they cannot place Gaussians in unobserved regions.
LaRa, despite being non-pixel-aligned, yields the worst FID due to its reliance on multi-view rendering loss alone.
While ReconViaGen also shows a good FID, it often produces misaligned predictions due to a high failure rate in its post-processing alignment.

\begin{table}[t]
\caption{\textbf{Ablation study.}
PSNR (dB) on GSO (1,030 objects) under full and partial observation, split into training-time and test-time ablations.}%
\label{table:ablations}
\centering
\small
\renewcommand{\arraystretch}{1.05}
\begin{tabular}{l@{\hspace{12pt}}l@{\hspace{12pt}}l}
\toprule
Configuration & Full (PSNR\up) & Partial (PSNR\up) \\
\midrule
Full method (Ours) & 
31.49& 
28.08\\
\midrule
\multicolumn{3}{l}{\textit{Training-time ablations}} \\
\quad No Flow Matching  & 28.40 \scriptsize{(\textcolor{red}{–3.09})} & 26.30 \scriptsize{(\textcolor{red}{–1.78})} \\
\quad No MV Rendering Loss & 29.66 \scriptsize{(\textcolor{red}{–1.83})} & 26.87 \scriptsize{(\textcolor{red}{–1.21})} \\
\quad No LoD Curriculum Training & 30.00 \scriptsize{(\textcolor{red}{–1.49})} & 26.42 \scriptsize{(\textcolor{red}{–1.66})} \\
\quad Without Patching (4K GS) & 30.47 \scriptsize{(\textcolor{red}{–1.02})} & 27.22 \scriptsize{(\textcolor{red}{–0.86})} \\
\quad Without Patching (8K GS) & 30.54 \scriptsize{(\textcolor{red}{–0.95})} & 26.75 \scriptsize{(\textcolor{red}{–1.33})} \\
\quad With Random Patching & 30.84 \scriptsize{(\textcolor{red}{–0.65})} & 27.16 \scriptsize{(\textcolor{red}{–0.92})} \\
\quad No Rendering Gradient Concatenation & 30.89 \scriptsize{(\textcolor{red}{–0.60})} & 27.51 \scriptsize{(\textcolor{red}{–0.57})} \\
\midrule
\multicolumn{3}{l}{\textit{Test-time ablations}} \\
\quad No Classifier-Free Guidance & 31.30 \scriptsize{(\textcolor{red}{–0.20})} & 27.50 \scriptsize{(\textcolor{red}{–0.60})} \\
\quad No Photometric Gradient Guidance & 31.10 \scriptsize{(\textcolor{red}{–0.40})} & 27.90 \scriptsize{(\textcolor{red}{–0.20})} \\
\bottomrule
\end{tabular}
\end{table}

\begin{figure}[th]
\centering
\includegraphics[trim=0cm 16.5cm 16.0cm 0.0cm, clip=true, width=1.0\linewidth]{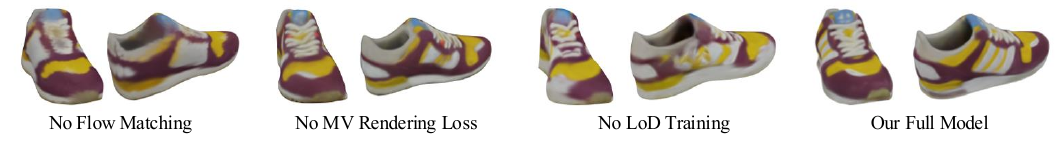}
\caption{\textbf{Qualitative Ablation results.} Using a generative model as the core predictor, a timestep-weighted rendering loss, and coarse-to-fine curriculum training prove the most significant contributors to the quality of our final results. }%
\label{fig:ablations}
\end{figure}

\subsection{Ablations}
\cref{table:ablations} ablates the key components of our method on GSO under both observation settings.
Removing flow matching, \ie replacing the DiT with a vanilla transformer and learnable latent tokens trained with photometric loss only (\cref{fig:pipeline}), causes the largest quality drop, confirming the importance of the generative formulation.
The timestep-weighted multi-view rendering loss (\cref{sec:recon_guided}) and coarse-to-fine curriculum training (\cref{sec:hier_patch}) are also significant; skipping the coarser 2K stage and training directly at 8K Gaussians degrades results. 
Removing patchification and processing all 8K Gaussians as individual tokens hurts performance because the doubled sequence length reduces training efficiency.\begin{wraptable}{r}{0.3\textwidth}
\vspace{-30pt}
\centering
\caption{\textbf{Number of input views.}
PSNR (dB) on GSO, full observations.}%
\label{table:num-views}
\small
\renewcommand{\arraystretch}{1.05}
\begin{tabular}{c@{\hspace{12pt}}c}
\toprule
\# Views & PSNR\up \\
\midrule
2 & 28.22 \\
4 & 31.49 \\
6 & 32.04 \\
9 & 32.39 \\
\bottomrule
\end{tabular}
\vspace{-16pt}
\end{wraptable}%
\unskip 
Conversely, halving the count to 4K Gaussians (matching our token budget but without patchification) also degrades quality due to reduced capacity. 
Patchification thus retains high local representational capacity while halving the sequence length. Replacing our LoD tree-based patchification with random groupings further degrades results, confirming that spatially coherent pairings provide a useful inductive bias. At each step we concatenate the gradient of the rendering loss with respect to the Gaussian parameters $\nabla_{\mathbf{z}} \mathcal{R}_{\text{Seen}}$ to the noisy inputs. 
Removing this step also reduces performance, showing that explicit gradient cues improve input reconstruction quality and input fidelity.

\inlineheading{Number of Input Views}
\cref{table:num-views} shows reconstruction quality as a function of the number of input views.
Increasing from two to four views provides the largest gain (+3.3\,dB), as the added viewpoints resolve geometric ambiguities.
Beyond this, returns diminish: six and nine views add +0.5\,dB and +0.9\,dB respectively, indicating that four views already cover most of the visible surface.
Nonetheless, our method consistently benefits from additional views when  available.

\section{Conclusion}%
\label{sec:conclusion}

We presented \emph{Free-Range Gaussians}, a non-grid-aligned approach to sparse-view 3D Gaussian reconstruction via flow matching.
Formulating reconstruction as conditional generation over Gaussian parameters overcomes the limitations of pixel-aligned methods (holes in unobserved regions), generate-and-reconstruct pipelines (dependence on synthesized-view placement), and volumetric methods (blurry conditional means from multi-view losses).
Our hierarchical LoD patching scheme groups sibling Gaussians into joint tokens for efficient training, and reconstruction-guided flow matching yields accurate results despite a moderate budget of Gaussians.
Experiments on Objaverse and Google Scanned Objects show consistent improvements over pixel-aligned (LGM \cite{TangCCWZL2024}, GS-LRM \cite{ZhangBTXZSX2024}) and voxel-based (LaRa \cite{ChenXETG2024}) methods, particularly under partial observation.

\inlineheading{Limitations and Future Work}
While our approach establishes a strong baseline, several avenues exist to further expand its capabilities.
Our current 8K Gaussian budget ensures tractable training but limits fine detail.
This can be addressed by adapting efficient attention mechanisms such as sparse \cite{Child2019GeneratingLS} or linear-complexity \cite{Katharopoulos2020TransformersAR} attention to scale to larger Gaussian counts, or by using our method as a coarse generator followed by an upsampling stage.
Our iterative denoising (50 steps) is slower than single-pass feed-forward methods; distillation \cite{SalimH2022} or flow straightening \cite{Liu2023InstaFlowOS} can reduce the step count while preserving quality.
Finally, operating directly in the 3D Gaussian parameter space opens avenues beyond reconstruction, including shape completion, semantic editing, and matching, bringing the versatility of 2D generative modeling into 3D.

\bibliographystyle{splncs04}
\bibliography{arxiv}

\end{document}